\documentclass{article}
\usepackage{spconf,amsmath,amssymb,graphicx,booktabs,array,float,balance}
\usepackage[
    colorlinks=true,
    linkcolor=blue,
    citecolor=blue,
    urlcolor=blue
]{hyperref}
\def\y{{\mathbf y}}
\def\h{{\mathbf h}}
\def\z{{\mathbf z}}
\def\D{{\cal D}}
\def\L{{\cal L}}

\title{WEATHERDIAGFLOW: EVIDENCE-GROUNDED RADAR NOWCASTING\\WITH DIAGNOSTIC FLOW REFINEMENT}

\name{Chunlei Shi$^{1}$, Yufeng Zhu$^{1}$, Yixiao Liang$^{2}$, Dan Niu$^{1}$, Yongchao Feng$^{3}$, Qiliang Wu$^{2,*}$, Jiong Wang$^{4,*}$}

\address{
\begin{tabular}{c}
\small $^{1}$Department of Automation, Southeast University, Nanjing, China\\[-0.2mm]
\small $^{2}$Beijing Fengyun Meteorological Science and Technology Development Co., Ltd., Beijing, China\\[-0.2mm]
\small $^{3}$State Key Laboratory of Virtual Reality Technology and Systems, Beihang University, Beijing, China\\[-0.2mm]
\small $^{4}$Department of Information Science and Technology, Fudan University, Shanghai, China
\end{tabular}
}

\begin{document}
\ninept
\maketitle

\begin{abstract}
\looseness=-1
Radar nowcasting is essential for short-term warning and emergency response, yet conventional systems mainly return future radar fields and provide limited support for operational communication and post-event verification. We formulate radar nowcasting as an evidence-grounded forecast--bulletin--audit task, in which a numerical forecaster produces both future radar fields and structured diagnostic evidence. Forecast-time bulletins use only model-available evidence, whereas post-event audits incorporate future radar truth only after the forecast horizon is observed. Based on this task formulation, WeatherDiagFlow predicts motion, growth and decay, heavy-echo risk, and uncertainty to condition rolling flow refinement, while frozen-scaffold residual calibration improves long-lead strong-echo preservation. A multi-agent layer converts the structured evidence into operational bulletins and independently generates verification audits without feeding textual outputs back into the forecaster. Experiments on FJRADAR demonstrate competitive overall performance and improved strong-echo event skill. WeatherDiagFlow therefore connects numerical prediction, evidence-grounded reporting, and auditable verification under a leakage-controlled protocol.
\end{abstract}


\begin{figure*}[t]
\centering
\includegraphics[width=0.96\textwidth]{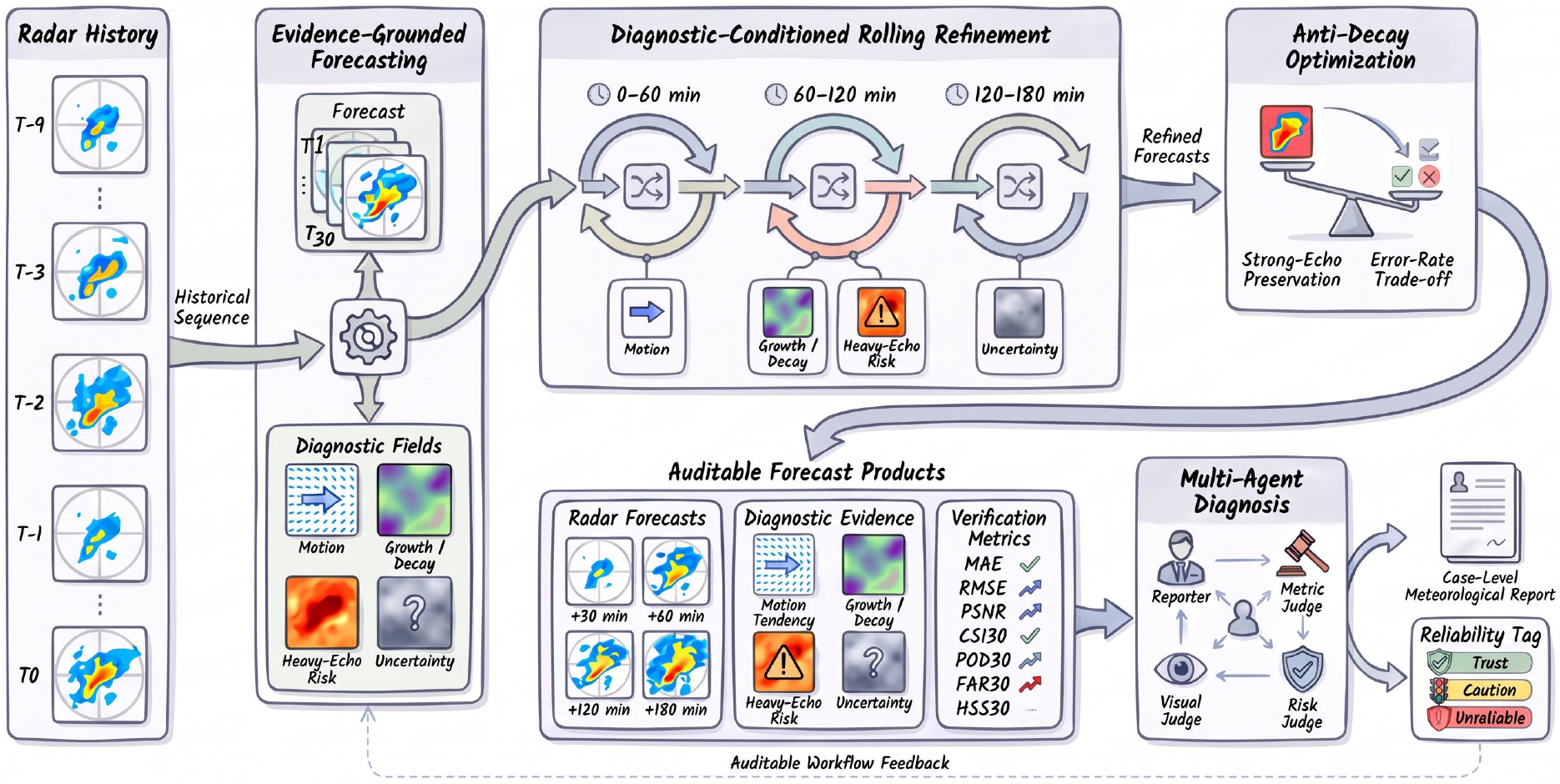}
\caption{WeatherDiagFlow separates forecast-time bulletin generation from post-event audit. Radar history and a coarse forecast produce predicted diagnostic fields that condition rolling flow refinement and a three-stage 0--3 hour weather bulletin. After future radar truth becomes available, verification metrics and error areas are added only to the audit product and reliability label. Textual agents package and check evidence but do not feed back into, or alter, the numerical forecast.}
\label{fig:framework}
\end{figure*}

\section{Introduction}
\label{sec:intro}

Accurate radar nowcasting supports convective warning, urban flood response, aviation weather services, and short-term emergency management. Radar echoes evolve through advection, growth, and decay, so a useful 0--3 hour forecast must retain coherent motion and localized high-reflectivity structures across lead times. Errors also accumulate across successive forecast chunks, making late-lead evolution especially vulnerable to spatial smoothing and weak-event bias. Yet mean absolute error (MAE) alone does not capture this requirement: smoothing uncertain futures can improve average error while removing the rare strong echoes most relevant to warning decisions and threshold-based emergency actions \cite{ravuri2021skillful,cao2023mir}.

Data-driven nowcasting has advanced from recurrent and convolutional predictors \cite{shi2015convlstm,ayzel2020rainnet,trebing2021smaat} to generative, residual-diffusion, cascaded, locality-aware, and motion-aware models \cite{ravuri2021skillful,yu2024diffcast,gong2024cascast,song2026exprecast,chen2026lmcm}. Recent work further explores pixel-space flow generation, structured radar representations, and physics-guided dynamic relations \cite{zhu2026pixelflowcast,shi2026wavec2r,chen2026veincast,camps2025artificial,reichstein2025early}. Despite these advances, most methods follow a prediction-centered formulation that maps radar history directly to future reflectivity fields. They often lose strong echoes as lead time increases and provide little structured evidence about displacement, growth or decay, heavy-echo risk, and predictive uncertainty. Thus, improving predictive skill alone is insufficient for an operationally complete nowcasting system; the forecast must also expose traceable evidence about what changes, where they occur, and how confidently they are predicted. The missing requirement is a structured account of forecast evolution that can be checked before and after issuance.

Meanwhile, multimodal models and large-language-model agents have been studied for weather-report generation and extreme-weather warning \cite{zheng2026weathersyn,ni2026siren,potter2021benefits}. However, numerical nowcasting, operational reporting, and post-event verification remain largely separate. Numerical models rarely expose structured evidence that can be directly interpreted by operational agents, whereas language-based systems are not inherently grounded in the internal diagnostics of radar predictors. Consequently, free-form reports may contain claims unsupported by numerical forecasts, while verification evidence derived from future observations must not enter previously issued guidance. An operational system therefore requires both a machine-readable evidence interface and a strict information boundary between forecast-time reporting and post-event auditing. This boundary links numerical prediction to accountable communication without allowing future truth to alter issued guidance.

To address these limitations, WeatherDiagFlow (WDF) unifies 0--3 hour radar forecasting, bulletin generation, and post-event auditing. WDF predicts motion, growth/decay, heavy-echo risk, and uncertainty to condition rolling flow refinement, while WDF-RC calibrates a frozen coarse scaffold to limit long-lead drift and preserve strong echoes.
At forecast issue time, agents convert only model-available evidence into 0--1, 1--2, and 2--3 hour bulletins \cite{jiang2025ewe,tang2026hvr,gonella2026crisitext}; after observations arrive, an independent audit evaluates errors and reliability. This boundary prevents future observations from entering issued guidance, while agents neither generate forecasts nor alter evaluation scores.
The main contributions are summarized as follows:
\begin{itemize}
\setlength{\itemsep}{0pt}
\setlength{\parskip}{0pt}
\setlength{\parsep}{0pt}
\setlength{\topsep}{1pt}
\item We formulate radar nowcasting as an evidence-grounded forecast--bulletin--audit task with an explicit boundary between forecast-time guidance and post-event verification.
\item We propose WDF with diagnostic-conditioned rolling flow refinement, where motion, growth/decay, heavy-echo risk, and uncertainty guide forecasting and operational interpretation. Its WDF-RC variant further preserves strong echoes through frozen-scaffold residual calibration.
\item Experiments on FJRADAR demonstrate competitive overall performance, improved strong-echo event skill, and evidence-consistent bulletin generation and auditing.
\end{itemize}

\section{Method}
\label{sec:method}

\subsection{Coarse future prior}

Given ten historical reflectivity frames $\h_{1:10}$ sampled every six minutes, the system predicts thirty future frames $\y_{1:30}$ over the next three hours. Reflectivity is normalized to $[0,1]$ during training and mapped back to dBZ for evaluation. The first module in Fig.~\ref{fig:framework} is a coarse forecaster,
\begin{equation}
    \tilde{\y}_{1:30}=F_{\rm c}(\h_{1:10}),
\end{equation}
which supplies a deterministic future scaffold rather than the final forecast. This scaffold provides a low-variance estimate of broad displacement and coverage, allowing later refinement to focus on local evolution instead of relearning the entire horizon. All subsequent modules use the valid-range mask $\Omega$ so losses and metrics exclude invalid radar pixels.

\subsection{Diagnostic field prediction}

The diagnostic encoder receives only the history and coarse forecast at inference time:
\begin{equation}
    \D_\phi(\h,\tilde{\y})=\{m,g_{1:30},r_{1:30},u_{1:30}\}.
\end{equation}
Here $m$ is a two-channel motion field, $g$ is echo growth/decay, $r$ is heavy-echo risk, and $u$ is an uncertainty proxy. During training, future truth defines supervision targets
\begin{equation}
    g_t^\star=\y_t-\y_{t-1},\quad
    r_t^\star=\mathbb{1}(\y_t\ge\tau),\quad
    u_t^\star=|\tilde{\y}_t-\y_t|,
\end{equation}
with $\tau$ set to the normalized risk threshold. Motion is constrained by brightness constancy and smoothness,
\begin{equation}
    \L_m=\|\h_{10}-{\cal W}(\h_{9},m)\|_{1,\Omega}
    +\beta\|\nabla m\|_{1,\Omega},
\end{equation}
where ${\cal W}$ denotes spatial warping. These ground-truth-derived targets supervise the diagnostic branch but are never supplied as forecast-time conditions.

\subsection{Rolling flow refinement}

The 30-frame horizon is partitioned into three one-hour chunks $\y^{(k)}=\y_{10(k-1)+1:10k}$. For chunk $k$, the rolling context $\h^{(k)}$ contains the original history plus previously generated chunks. A pixel-space flow model estimates
\begin{equation}
    v_\theta=v_\theta(\z_t,\h^{(k)},\tilde{\y}^{(k)},m,g^{(k)},r^{(k)},u^{(k)},t),
\end{equation}
where $\z_t=(1-t)\epsilon+t\y^{(k)}$ interpolates Gaussian noise $\epsilon$ and the target chunk. The flow target is $\y^{(k)}-\epsilon$, and the complete training objective is given by
\begin{equation}
\begin{split}
\L={}&\L_{\rm vel}+\lambda_{\rm clean}\L_{\rm clean}
+\lambda_{\rm echo}\L_{\rm echo}+\lambda_c\L_{\rm coarse}\\
&+\lambda_g\L_g+\lambda_r\L_r+\lambda_u\L_u+\lambda_m\L_m .
\end{split}
\end{equation}
The stage-1 diagnostic-flow weights are
\begin{equation}
\begin{gathered}
(\lambda_{\rm clean},\lambda_{\rm echo},\lambda_c,\lambda_g)
=(0.5,0.8,0.08,0.2),\\
(\lambda_r,\lambda_u,\lambda_m,\beta,\lambda_{\rm fa},\lambda_{\rm miss})
=(0.2,0.1,0.05,0.01,0.5,0.2).
\end{gathered}
\end{equation}
Late-chunk weighting increases from 1.0 to 1.5, and inference uses eight Euler steps per chunk. The diagnostic-flow forecaster has 4.114M parameters versus 3.663M for the matched rolling pMF baseline.

To prevent the refinement process from amplifying errors in its own earlier outputs, we freeze the coarse diagnostic-flow scaffold and learn a residual correction around it. Instead of transporting noise directly to the complete target chunk, the residual branch learns
\begin{equation}
    \mathbf r^{(k)}=\y^{(k)}-\tilde{\y}^{(k)},\qquad
    \hat{\y}^{(k)}=\operatorname{clip}\!\left(\tilde{\y}^{(k)}+\hat{\mathbf r}^{(k)},0,1\right).
\end{equation}
The residual path starts from low-amplitude noise, which restricts the correction to deviations that the scaffold does not explain. It is supplemented by soft echo-area and echo-mass matching, temporal-tendency consistency, persistent-echo protection where both the coarse forecast and truth contain echoes, and mixed scaffold/generated histories during training, with generated histories used at inference. These constraints target the main failure modes of long-horizon refinement: excessive area growth, disappearance of persistent echoes, and feedback from an inaccurate generated history.

Positive and negative residuals are scaled separately, and the history passed to the next rolling chunk is anchored to the coarse forecast:
\begin{equation}
\begin{split}
\bar{\mathbf r}^{(k)}={}&\alpha_{+}[\hat{\mathbf r}^{(k)}]_{+}-\alpha_{-}[-\hat{\mathbf r}^{(k)}]_{+},\\
\hat{\y}^{(k)}_{\rm RC}={}&\operatorname{clip}\!\left(\tilde{\y}^{(k)}+\bar{\mathbf r}^{(k)},0,1\right),\\
\h^{(k+1)}={}&\rho\hat{\y}^{(k)}_{\rm RC}+(1-\rho)\tilde{\y}^{(k)}.
\end{split}
\end{equation}
We use $(\alpha_{+},\alpha_{-},\rho)=(0.60,1.00,0.60)$. Separating positive and negative corrections limits systematic echo amplification, while anchoring the next context to the scaffold reduces autoregressive feedback and preserves broad-scale evolution.

The anti-decay extension keeps the frozen scaffold and residual path but increases the training pressure against late-lead echo decay. Its objective adds anti-decay terms to the residual loss,
\begin{equation}
\begin{split}
    \L_{\rm AD}={}&\L_{\rm RC}
    +\lambda_{\rm miss}\L_{\rm miss}
    +\lambda_{\rm csi}\L_{\rm softCSI}\\
    &+\lambda_{\rm area}\L_{\rm area}
    +\lambda_{\rm mass}\L_{\rm mass}
    +\lambda_{\rm late}\L_{\rm late},
\end{split}
\end{equation}
where the added terms penalize missed echoes, encourage differentiable threshold skill, preserve echo area and echo mass, and upweight later forecast chunks. The resulting residual-calibration and anti-decay variants are compared in the ablation study, with the preference-calibrated variant selected using validation data before the independent test evaluation.

\subsection{Forecast--bulletin--audit agents}

The agent layer converts numerical forecasts into three evidence-grounded products without feeding text back to the forecaster. The first product is a machine-readable evidence record. At forecast issue time, it is limited to model-available quantities:
\begin{equation}
    e_{\rm ft}=\{m,g,r,u\}.
\end{equation}
After future radar truth arrives, verification augments this into
\begin{equation}
    e_{\rm audit}=\{e_{\rm ft},q,a_{\rm miss},a_{\rm fa},o_{\rm ue}\},
\end{equation}
where $q$ contains coarse/refined MAE and their difference, $a_{\rm miss}$ and $a_{\rm fa}$ are missed-echo and false-alarm area fractions, and $o_{\rm ue}$ is uncertainty--error overlap. The second product is a forecast-time bulletin $B$ generated only from $e_{\rm ft}$; it describes overall echo motion, 0--1/1--2/2--3 hour evolution, attention regions, and evidence-bounded confidence. The third product is a post-event audit $A$ generated from $e_{\rm audit}$ after truth arrives.

Five role-specific agents process the products:
\begin{equation}
\begin{aligned}
(B,A)&=A_{\rm rep}(e_{\rm ft},e_{\rm audit}),\\
s_m&=A_{\rm met}(B,A,e),\quad s_v=A_{\rm vis}(B,I),\\
\mathbf{s}_q&=A_{\rm qual}(B,e_{\rm ft}),\\
\ell&=A_{\rm risk}(A,e_{\rm audit},s_m,s_v,\mathbf{s}_q).
\end{aligned}
\end{equation}
Here $s_m$ and $s_v$ are metric and visual consistency scores, $I$ denotes forecast figures, and $\mathbf{s}_q$ contains evidence consistency, temporal coherence, spatial correctness, meteorological completeness, and operational usefulness. The metric judge additionally rejects future-truth leakage into $B$, while $\ell\in\{\mathrm{trust},\mathrm{caution},\mathrm{unreliable}\}$ is assigned only to the completed post-event audit. Deterministic fallbacks reproduce the same product schema when an external VLM/LLM service is unavailable.

\section{Experiments}
\label{sec:experiments}

\subsection{Dataset and protocol}

\noindent\textbf{Dataset.} We use FJRADAR composite reflectivity from the QZSFZ radar around Fuzhou ($118.160^\circ$--$121.155^\circ$E, $24.567^\circ$--$27.262^\circ$N). Each $300\times270$ sample contains ten historical frames and thirty future frames at six-minute spacing. We evaluate an event-focused precipitation subset, retaining cases where any future frame has at least 5\% valid pixels above 10 dBZ. The chronological train/validation/test split contains 30,665/6,345/2,125 samples from 5 Jan. 2022--10 Jul. 2023, 16 Jul.--17 Sep. 2023, and 23 Sep.--29 Nov. 2023.

\noindent\textbf{Protocol.} All methods share the 2,125-case test manifest. Metrics are computed over the valid mask by pooling pixels across all test cases and all 30 future frames; no-positive frames are kept. The compact table emphasizes event-oriented and perceptual scores: valid-frame-averaged SSIM and LPIPS, globally pooled CSI10/20/30, POD20/30, and HSS20/30. MAE, PSNR, FAR30, and 40-dBZ scores are omitted from the compact table because they are less aligned with the strong-echo retention claim or are too sparse. Baselines include SmaAt-UNet, rolling pMF conditioned on the same SmaAt coarse forecast, residual-diffusion DiffCast \cite{yu2024diffcast}, deterministic CasCast/EarthFormer \cite{gong2024cascast}, and exPreCast \cite{song2026exprecast} under the same ten-input/thirty-output evaluation protocol for all runs.

\begin{table*}[!t]
\centering
\caption{Event-oriented and perceptual nowcasting results on the common 2,125-case FJRADAR test set. WDF-RC-AD (Ours) denotes the validation-selected preference-calibrated anti-decay variant.}
\label{tab:main}
\scriptsize
\setlength{\tabcolsep}{1.7pt}
\renewcommand{\arraystretch}{1.06}
\begin{tabular*}{\textwidth}{@{\extracolsep{\fill}}l*{9}{c}@{}}
\toprule
Method & SSIM$\uparrow$ & LPIPS$\downarrow$ & CSI10$\uparrow$ & CSI20$\uparrow$ & CSI30$\uparrow$ & POD20$\uparrow$ & POD30$\uparrow$ & HSS20$\uparrow$ & HSS30$\uparrow$ \\
\midrule
SmaAt-UNet~\cite{trebing2021smaat} & 0.3283 & \underline{0.4040} & \underline{0.2901} & 0.2004 & 0.0608 & 0.2374 & 0.0666 & 0.3154 & 0.1128 \\
Rolling pMF + coarse~\cite{lipman2023flow} & 0.2108 & 0.4587 & 0.1210 & 0.0638 & 0.0169 & 0.0679 & 0.0181 & 0.1104 & 0.0320 \\
DiffCast~\cite{yu2024diffcast} & \underline{0.3538} & 0.4401 & 0.2566 & 0.1565 & 0.0612 & 0.2123 & 0.0756 & 0.2458 & 0.1120 \\
CasCast~\cite{gong2024cascast} & 0.3318 & 0.4050 & \textbf{0.3110} & \underline{0.2322} & 0.0894 & \underline{0.2929} & 0.1060 & \underline{0.3560} & 0.1611 \\
exPreCast~\cite{song2026exprecast} & 0.3018 & 0.4118 & 0.2773 & 0.1783 & 0.0485 & 0.2068 & 0.0517 & 0.2852 & 0.0910 \\
\midrule
WDF-RC-AD (Ours) & \textbf{0.3730} & \textbf{0.2146} & 0.2883 & \textbf{0.2448} & \textbf{0.1351} & \textbf{0.3811} & \textbf{0.3384} & \textbf{0.3757} & \textbf{0.2273} \\
\bottomrule
\end{tabular*}
\end{table*}

\subsection{Quantitative nowcasting results}

\begin{table}[!t]
\centering
\caption{Diagnostic-field checks for the WeatherDiagFlow diagnostic branch.}
\label{tab:diagnostics}
\scriptsize
\setlength{\tabcolsep}{2.5pt}
\renewcommand{\arraystretch}{1.02}
\begin{tabular}{lccccc}
\toprule
Motion & Growth & Risk & Risk & Risk & U--err. \\
resid. & MAE & CSI & POD & FAR & corr. \\
\midrule
0.0151 & 2.2935 & 0.2078 & 0.2603 & 0.3624 & 0.3706 \\
\bottomrule
\end{tabular}
\end{table}

\begin{figure}[!t]
\centering
\includegraphics[width=0.8\columnwidth,keepaspectratio]{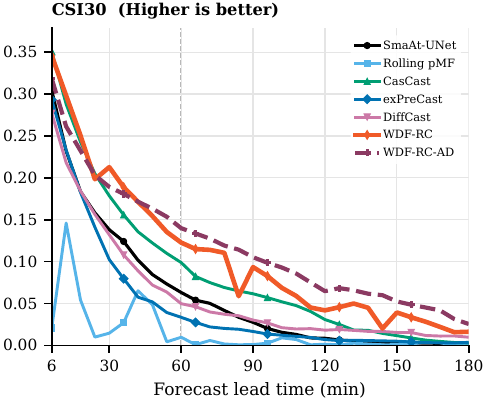}
\caption{Lead-time CSI30 on the common FJRADAR test set, pooled over the valid pixels and all forecast cases at each lead time. The 30-dBZ threshold emphasizes strong-echo event detection, while the 0--180 min curves show how skill changes as temporal uncertainty accumulates. The comparison includes the diagnostic-flow baselines and the retained WDF-RC and WDF-RC-AD variants.}
\label{fig:lead-time-main}
\end{figure}

Figure~\ref{fig:lead-time-main} shows that the retained WeatherDiagFlow variants improve strong-echo event skill over the forecast horizon: CSI30 rises from 0.0169 for matched rolling pMF to 0.1284 for WDF-RC and 0.1351 for WDF-RC-AD. The corresponding POD30 values are 0.3005 and 0.3384, respectively, while the calibrated WDF-RC-AD has FAR30=0.8164.
Pixel-wise losses favor conservative fields under uncertain convective evolution, whereas thresholded scores reward retained high-reflectivity cores even when displaced. We therefore report WDF-RC-AD as the primary retained variant and use WDF-RC as the residual-calibration reference in the ablation analysis.
Table~\ref{tab:main} reports globally pooled event-oriented test results. WDF-RC-AD gives the best SSIM, LPIPS, CSI20/30, POD20/30, and HSS20/30 among the retained methods, while CasCast remains strongest at CSI10 and competitive at CSI20.
Table~\ref{tab:diagnostics} provides a sanity check that the diagnostic branch carries nontrivial forecast-time signal. The heavy-echo risk map reaches 0.2078 CSI at the 10-dBZ risk-supervision threshold, and uncertainty has a positive 0.3706 correlation with realized absolute error. These checks support the use of diagnostic fields as conditioning evidence, but they are not a full calibration study: motion is evaluated through a brightness-constancy residual rather than ground-truth optical flow, and event-level calibration remains future work.

\subsection{Case-level and operational diagnosis}

\begin{figure}[!t]
\centering
\includegraphics[width=\columnwidth,keepaspectratio]{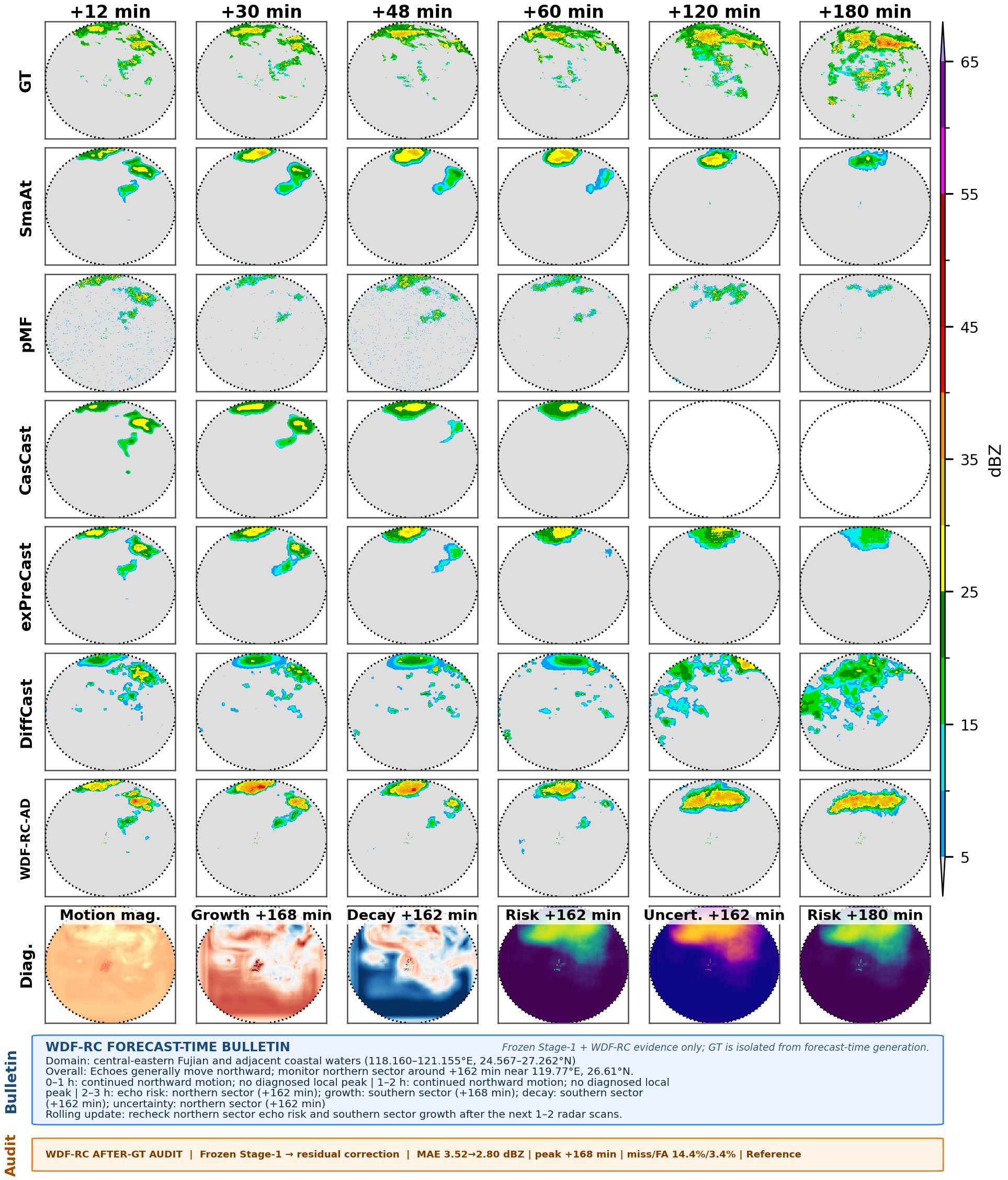}
\caption{Representative large-area case with forecast-time evidence and post-event audit separated.}
\label{fig:case}
\end{figure}

Figure~\ref{fig:case} shows one event at six lead times over 0--3 h. WDF-RC-AD retains strong-echo structure while limiting late-lead area inflation. Forecast-time products exclude GT information; verification evidence is used only by the post-event audit.

We construct a 32-case task set from held-out cases. Forecast-time bulletins use only available evidence, whereas audits use future-derived errors and areas. Report quality is scored for evidence consistency, temporal coherence, spatial correctness, completeness, and operational usefulness; these checks verify the task boundary rather than human preference. This separation makes the numerical forecast, issued guidance, and post-event assessment independently traceable.

\subsection{Ablation study}

Table~\ref{tab:ablation} compares WDF, residual calibration, and the preference-calibrated AD extension. Residual calibration raises CSI30/POD30 from 0.0274/0.0298 to 0.1284/0.3005, indicating that the frozen scaffold provides a more stable basis for strong-echo refinement. AD further reaches 0.1351/0.3384 with HSS30=0.2273 and FAR30=0.8164, close to the RC reference of 0.8169. The result supports anti-decay calibration as a threshold-skill adjustment rather than unconstrained intensity amplification, while preserving the evidence interface used by the downstream bulletin and audit products.

\begin{table}[!h]
\centering
\caption{Component ablation on the common FJRADAR test manifest. D, RC, and AD denote diagnostic rolling flow, residual calibration, and anti-decay preference calibration, respectively. Best completed results are in \textbf{bold}; FAR is lower-is-better.}
\label{tab:ablation}
\scriptsize
\setlength{\tabcolsep}{1.2pt}
\renewcommand{\arraystretch}{1.05}
\begin{tabular}{@{}lccc|cccccc@{}}
\toprule
Variant
& D
& RC
& AD
& CSI10$\uparrow$
& CSI20$\uparrow$
& CSI30$\uparrow$
& POD30$\uparrow$
& FAR30$\downarrow$
& HSS30$\uparrow$ \\
\midrule
WDF
& $\checkmark$
& $-$
& $-$
& 0.0645
& 0.0471
& 0.0274
& 0.0298
& \textbf{0.7401}
& 0.0519 \\

WDF-RC
& $\checkmark$
& $\checkmark$
& $-$
& 0.2599
& 0.2266
& 0.1284
& 0.3005
& 0.8169
& 0.2200 \\

WDF-RC-AD
& $\checkmark$
& $\checkmark$
& $\checkmark$
& \textbf{0.2883}
& \textbf{0.2448}
& \textbf{0.1351}
& \textbf{0.3384}
& \underline{0.8164}
& \textbf{0.2273} \\
\bottomrule
\end{tabular}
\end{table}

\section{Conclusion}
\label{sec:conclusion}

We introduced WeatherDiagFlow, an evidence-grounded radar nowcasting framework that formulates 0--3 hour prediction as a forecast--bulletin--audit task. WDF uses predicted motion, echo growth/decay, heavy-echo risk, and uncertainty to condition rolling flow refinement, while WDF-RC adds frozen-scaffold residual calibration for stable long-lead prediction. The explicit evidence boundary keeps future observations out of forecast-time bulletins and reserves them for post-event audits. Future work will finalize FAR-constrained anti-decay selection and evaluate diagnostic calibration and bulletin usefulness across broader events.

\clearpage

\section{Compliance with Ethical Standards}

This study uses radar reflectivity data and automated model-based evaluation. It does not involve human participants or animals; therefore, ethical approval was not required.
\par\nopagebreak[4]

\flushbottom
\balance
\bibliographystyle{IEEEbib}
\bibliography{refs}

\end{document}